\documentclass{article}

\usepackage{arxiv}

\usepackage[utf8]{inputenc} 
\usepackage[T1]{fontenc}    
\usepackage{hyperref}       
\usepackage{url}            
\usepackage{booktabs}       
\usepackage{amsfonts}       
\usepackage{nicefrac}       
\usepackage{microtype}      
\usepackage{lipsum}
\usepackage{graphicx}
\usepackage{amsmath}
\usepackage{amssymb}
\usepackage[utf8]{inputenc} 
\usepackage[T1]{fontenc}    
\usepackage{amsfonts}       
\usepackage{xcolor}         
\usepackage{cuted}
\usepackage{fancyvrb}
\usepackage{multirow}

\title{Inf-CP: A Reliable Channel Pruning based on Channel Influence}

\author{
 Bilan Lai \\
  School of Artificial Intelligence \\
  of Nanjing University \\
  \texttt{mg1933032@smail.nju.edu.cn} \\
   \And
 Haoran Xiang \\
  School of Artificial Intelligence \\
  of Nanjing University \\
  \texttt{mg20370040@smail.nju.edu.cn} \\
  \And
 Furao Shen \\
  School of Artificial Intelligence \\
  of Nanjing University \\
  \texttt{frshen@nju.edu.cn} \\
}

\begin{document}
\maketitle
\begin{abstract}
One of the most effective methods of channel pruning is to trim on the basis of the importance of each neuron. However, measuring the importance of each neuron is an NP-hard problem. Previous works have proposed to trim by considering the statistics of a single layer or a plurality of successive layers of neurons. These works cannot eliminate the influence of different data on the model in the reconstruction error, and currently, there is no work to prove that the absolute values of the parameters can be directly used as the basis for judging the importance of the weights. A more reasonable approach is to eliminate the difference between batch data that accurately measures the weight of influence. In this paper, we propose to use ensemble learning to train a model for different batches of data and use the influence function (a classic technique from robust statistics) to learn the algorithm to track the model’s prediction and return its training parameter gradient, so that we can determine the responsibility for each parameter, which we call "influence", in the prediction process. In addition, we theoretically prove that the back-propagation of the deep network is a first-order Taylor approximation of the influence function of the weights. We perform extensive experiments to prove that pruning based on the influence function using the idea of ensemble learning will be much more effective than just focusing on error reconstruction. Experiments on CIFAR shows that the influence pruning achieves the state-of-the-art result. 
\end{abstract}


\section{Introduction}
\label{sec:intro}

In the deep network, due to the network structure having a large number of levels and nodes, this brings enormous memory load and computing power consumption. Most of the marginalized devices cannot support intensive task calculations, which largely hinders the commercialization of deep learning methods. Therefore, it is necessary to consider reducing the amount of calculation and memory load through model compression. Pruning is an important category in model compression. It can be divided into structured pruning and unstructured pruning. Unstructured pruning \cite{lecun1990optimal,han2015learning,ding2019global} requires additional hardware and library support, or the speed cannot be increased \cite{han2016eie}. For structured pruning \cite{huang2018data,liu2017learning,zhuang2020neuron}, if the pruning unit is too large (e.g., network layer, network group), it will cause a great loss of accuracy. In order to ensure as much accuracy as possible and to ensure that it can run on ordinary CPU/GPU devices, our pruning work is neuron-level pruning.

For neuron-level pruning, one of the best methods is to get the importance of each neuron and cut out the unimportant channels by sorting the importance. However, directly measuring the importance of neuron weight is an Np-hard problem. Therefore, single-layer or continuous multi-layer information is usually collected for pruning (e.g., reconstruction error), but reconstruction error can only be a way of indirect approximate expression of influence \cite{luo2017thinet,he2017channel,yu2018nisp,lee2019snip}, and because these tasks cannot eliminate the influence of different data on reconstruction errors, Therefore, there are still some problems with this expression. \cite{koh2020understanding} proposes to use the influence function to test the influence of the data entering the black-box model. And \cite{koh2019accuracy} proved that in many different types of groups, for a series of actual data sets, even if the absolute and relative errors are zero, the influence predicted by the influence function of the group data is close to the actual influence because the weight influence reflected by the reconstruction error is not convincing. Inspired by the theories of \cite{koh2019accuracy} and \cite{koh2020understanding}, we proposed a method for measuring the influence of weights based on the influence function (a classic technique from robust statistics). At the same time, we adopted a technique similar to ensemble learning (Bagging \cite{breiman1996bagging}), dividing the data set into multiple batches, and performing generalized learning on the importance of the same neuron measured in each batch of data, so as to eliminate the impact of measurement effects caused by different data.

In addition, in order to make the pruning strategy and the process of fine-tuning the model promote each other, we combined the common training method. Filter selection and model fine-tuning have been integrated into a single end-to-end trainable framework. \cite{luo2020autopruner} proves that the two processing steps can promote each other. The combined use of the two steps will give full play to their advantages, which can significantly save training time.

To summarize our contributions as follows: 
\begin{itemize}
    \item We proved that the weight gradient of the backpropagation of the model is the first-order Taylor approximation of the influence function of the weight. In practice, this makes weight influence easier to measure.
    \item Based on the idea of ensemble learning, we eliminate the influence strategy error caused by different batches of data. Precisely, the weight influence is measured on batch data, and the influence is generalized learning using a convolutional layer. Therefore, through the predictive decision strategy of the convolutional layer, the effect caused by different batches of data can be eliminated.
    \item Our work has the ability to automatically adjust to the actual pruning work. We add a regular encouragement to the loss function to encourage the model pruning strategy to move closer to its target strategy. This prompted model to take into account the accuracy of the model and pruning strategies to achieve automatic adjustment of the pruning strategy.
    \item Compared with the previous state-of-the-art algorithms, our method is easy to implement. This method has proved its effectiveness on VGG and ResNet, and it has better performance on CIFAR.
\end{itemize}

\section{Pruning based on Influence}
\label{gen_inst}
In this section, we focus on our intuitive approach and introduce its implementation details. To simplify the description, we only discuss the convolutional layer and the fully connected layer. In fact, our pruning method can also be applied to some other layer types, as long as their basic mathematical operations can be derived.

\subsection{Notation}
First of all, We clarify the concepts in this article. It is assumed that the weight of the DNN model can be expressed as $\{\mathbf{W}_l^{k}: 0 \leq k\leq C_l, 0 \leq l \leq N \}$, in which $\mathbf{W}_l^k$ represents the connection weight matrix of the $k$-th filter in the $l$-th layer, $C_l$ represents the number of channels in the $l$-th layer, and $N$ represents the number of layers of the DNN model.

For a convolutional layer with a $p$-dimensional input and a $q$-dimensional output, the number of weighted convolutions with $W\times H$ dimensions is $q^k \times p^k \times W \times H$. For fully connected layers with $p$-dimensional inputs and $q$-dimensional outputs, the number of weights is $q^k\times p^k$.

In order to represent the sparse vector where a part of the neuron connection is deleted, we use $\{ T_l: 0\leq l\leq N\}$.
Tl is a binary vector whose entries indicate the connection status of the first layer network channel, that is, whether they are currently pruned, and $T_l$ can be regarded as the mask vector of the network layer channel.

\subsection{Weight Influence}

Since our goal is to prune the network channel, we should learn the required sparse model from the influence of the convolution parameters in the channel. 
Obviously, the key is to abandon the unimportant parameters and keep the important ones. This is an NP-hard problem, but our work can find an approximate solution.

First of all, we set $\mathbf{W}^{k,(i,j)}_l$ as the parameter weight of the $i$-th row and $j$-th column in the $l$-th layer and $k$-th convolution kernel. Taking the $l$-th layer as an example, we suggest solving the following optimization problems. We propose the loss function of the DNN model: 
\begin{equation}
J\left(\mathbf{W};\mathcal{D}\right)=L\left( \mathbf{1} \odot \mathbf{W}_l; \mathcal{D}\right),
\label{eq1}
\end{equation}
in which $\odot$ represents the Hadamard product, $\mathbf{W}_l$ represents the weight of the $l$-th layer, and $\mathbf{1}$ represents an all-one matrix of the same size as $\mathbf{W}_l$. After removing the single weight $\mathbf{W}^{k,(i,j)}_l$ in the $k$-th convolution kernel of the $l$-th layer, the loss function of the DNN model:
\begin{equation}
J\left(\mathbf{W};\mathcal{D}\right)=L\left( (\mathbf{C}^{k,(i,j)} )\odot \mathbf{W}_l; \mathcal{D}\right),
\label{eq2}
\end{equation}
in which Where $L(\cdot)$ is the loss function, and $\mathbf{C}$ represents the binary operation matrix corresponding to a certain weight in the current layer of the convolution kernel. We use 0 and 1 to indicate the pruning state of a single weight of the convolution kernel, 0 means pruning, and 1 means reserved. In other words, $\mathbf{C}^{k,(i,j)}$ is a binary matrix with indexs $(k, i, j)$ whose value is 0 and the rest are 1, which means that the weight of the $i$-th row and the $j$-th column in the $k$-th channel is clipped. 
 For the weight $\mathbf{W}^{k,(i,j)}_l$, the loss difference caused by retaining and pruning the weight can be directly used as the influence $\mathcal{I}$ of the weight:
 \begin{equation}
    \mathcal{I}^k_l = \Delta L_{l}^k(\mathbf{w}_l^k ; \mathcal{D})
\label{eq3}
\end{equation}
 which we call the "weight Influence" of $\mathbf{W}^{k,(i,j)}_l$ to the model, and $\mathcal{I}_l^k$ is the weight influence of the $k$-th channel in the $l$-th layer. Then, the loss difference for the deleted weight $\mathbf{W}^{k,(i,j)}_l$ is as equation \ref{eq4}.
\begin{figure*}[h]
\begin{equation}
    \Delta L_{l}^{k}(\mathbf{W}_l^{k} ; \mathcal{D})=\sum_i^{W} \sum_j^{H}\left[ L \left(\mathbf{1} \odot \mathbf{w}_l^{k,(i,j)} ; \mathcal{D}\right)-L\left(\mathbf{C}^{k,(i,j)} \odot \mathbf{w}_l^{k,(i,j)} ; \mathcal{D}\right)\right],
\label{eq4}
\end{equation}
\end{figure*}
$W$ and $H$ in it represent the width and height of the convolution kernel, respectively. It can be seen from \cite{koh2019accuracy} that the influence measurement of a single weight is not directly related to the influence measurement of other weights, so the loss difference of a single weight in the channel can be directly accumulated into equation \ref{eq4}. However, $\mathbf{C}$ in equation \ref{eq4} is a binary matrix, and the binary system makes equation \ref{eq4} unable to differentiate and derive. We suggest changing the equation \ref{eq4} to equation \ref{eq5}.
\begin{figure*}[h]
\begin{equation}
     \Delta \hat{L}_{l}^{k}(\mathbf{W}_l^{k} ; \mathcal{D})=\sum_i^{W} \sum_j^{H}\left[ L \left(\mathbf{1} \odot \mathbf{w}_l^{k,(i,j)} ; \mathcal{D}\right)-L\left(\alpha \mathbf{C}^{k,(i,j)} \odot \mathbf{w}_l^{k,(i,j)} ; \mathcal{D}\right)\right],0\leq \alpha \leq 1,
\label{eq5}
\end{equation}
\begin{equation}
    \mathcal{I}^k_l = \Delta L_{l}^k(\mathbf{w}_l^k ; \mathcal{D}) \approx \frac{\partial L(\mathbf{C} \odot \mathbf{w}_l^k ; \mathcal{D})}{\partial \mathbf{C}},
\end{equation}
\begin{equation}
    \frac{\partial L(\mathbf{C} \odot \mathbf{w}_l^k ; \mathcal{D})}{\partial \mathbf{C}}=\lim _{\alpha \rightarrow 0}\sum_i^W \sum_j^H \frac{L(\mathbf{1} \odot \mathbf{w}^{k,(i,j)}_l ; \mathcal{D})-L\left(\alpha\mathbf{C}^{k,(i,j)}\odot \mathbf{w}^{k,(i,j)}_l ; \mathcal{D}\right)}{\alpha \mathbf{C}^{k,(i,j)}}.
\label{eq6}
\end{equation}
\end{figure*}
We suggest adding $\alpha$ times perturbation to a single weight $\mathbf{W}^{k,(i,j)}_l$, and deriving the perturbation factor $\alpha$ to measure the derivative of the weight $\mathbf{W}^{k,(i,j)}_l$. It can be seen from \cite{koh2020understanding} that the derivative value of $\alpha$ can be expressed as the influence of a single weight $\mathbf{W}^{k,(i,j)}_l$. We assume that by accumulating the influence of all weight measurements in the same convolution kernel, the weight influence of the current channel can be obtained. We use the weight influence of the measured channel as the decision index of the pruning strategy. We set the $\alpha$ perturbation derivative for a single weight euation \ref{eq6}.
From formula \ref{eq6}, we obtain  the influence  of the weight $\mathbf{W}^k_l$ by deriving $\alpha$. Equation \ref{eq5} can be used to determine the influence of $\mathbf{W}^k_l$ by using backpropagation, which will be described in detail in \ref{backpro}.
\subsection{Backpropagation}
\label{backpro}
\begin{figure*}
  \centering
  \includegraphics[height=7cm, width=13cm]{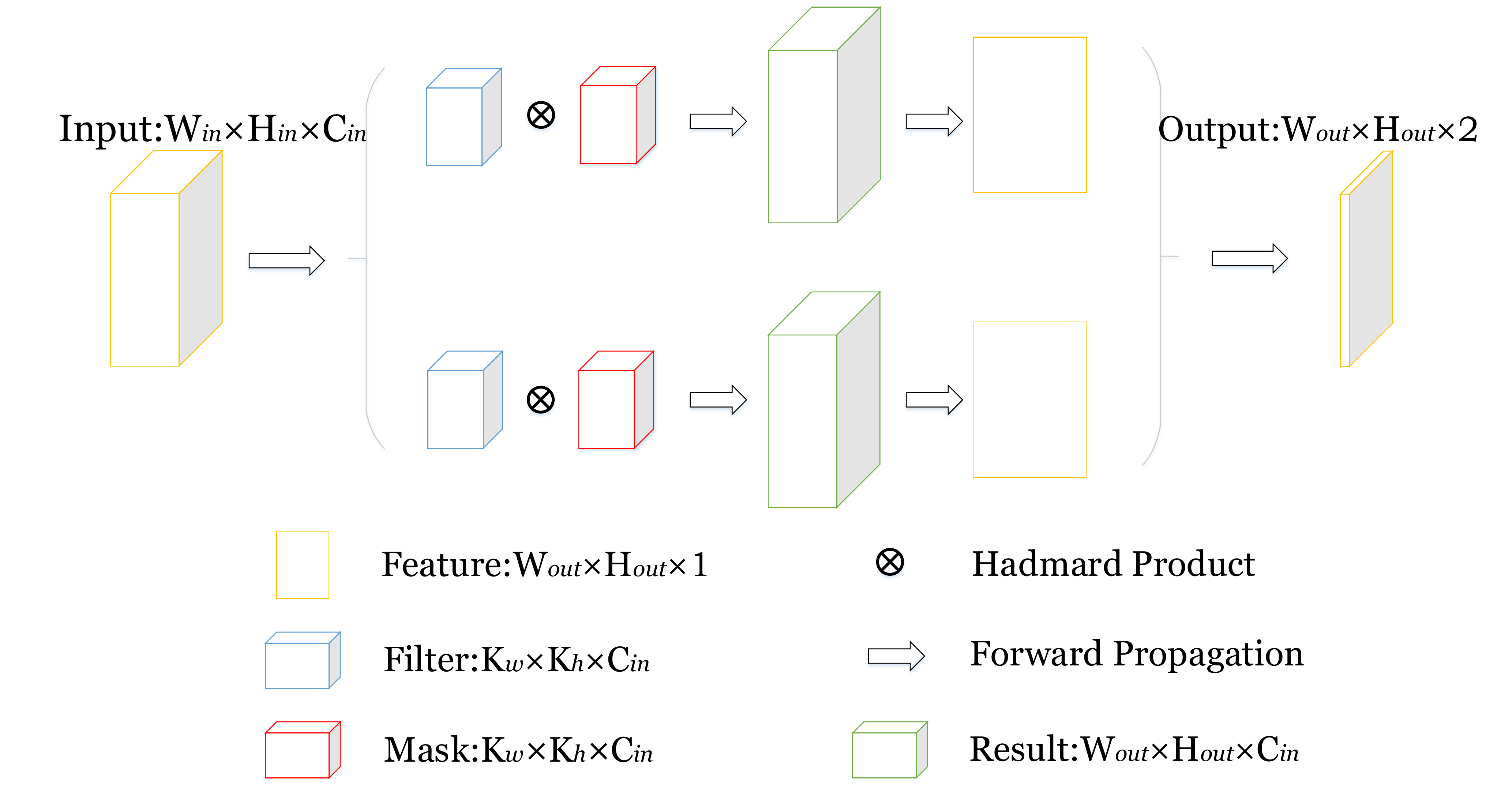}
  \caption{Input through the Filter of 2 channels to get the result. Then accumulate the Result into a two-dimensional Feature in the channel. Finally, the Features of each channel are spliced to get the Output.The Mask represents a filter of the same size as the channel filter, and is initialized to a constant value of 1.}
\label{fg1}
\end{figure*}

Since the influence $ \mathcal{I}^k_l$ can be determined by the equation \ref{eq6}, we only need to use backpropagation to obtain the derivative result of $\alpha$ as the weight influence. In order that sensitivity can be further used as a pruning strategy, we have formulated a processing scheme for influence $ \mathcal{I}^k_l$ as shown in Figure \ref{fg1}.
We add an auxiliary filter of the same size as the filter, which is initialized to a constant 1, called "Mask" filter, which combines the two filters in the channel into one filter by the Hadmard product. At the same time, we control the original filter in the reverse direction. 
The gradient is not saved during propagation, but the gradient value of the Mask filter is saved. 

For a single weight $M^{k,(i,j)}$ of the Mask Filter, due to the combined effect of the Hadmard product, the Mask filter is equivalent to converting each weight of the original filter to $M^{k,(i,j)}$ times. 
In the equation \ref{eq6}, $M^{k,(i,j)}$ is equivalent to the $\alpha$ corresponding to a single weight of the original filter, which means that the gradient of the Mask filter can reflect the importance of the gradient of the filter.

In order to make the gradient calculated by the Mask filter have generalization, we added a layer of convolution to process the gradient of the Mask filter, and the obtained result was binarized. Here we use the $sigmoid$ function to complete the binarization, which is mapped to a probability value of 0-1, which is the basis for whether the current channel is pruned. 
The importance determination process simplifies the original NP-hard problem. 
For the sake of brevity of the topic, we will discuss the function $sigmoid(\cdot)$ in Section \ref{binary}.
\subsection{Training}
\label{train}
During the training phase, we use the current mini-batch samples as input to calculate the gradient of the Mask filter through backpropagation, and use Equation \ref{eq6} to obtain the influence of the entire channel and a single weight in the channel. 

On the one hand, we choose to delete the channels whose channel influence is less than the threshold, and generate the channel target pruning strategy vector. We use binary vector $T_l$ representation, the value of the vector is represented by 0-1, and each value indicates whether the corresponding channel is pruned, and if it is 1, the channel is reserved. 
On the other hand, we pass the influence of a single weight in all channels of the current layer of the model through a layer of micro-convolutional layer $K$ to obtain the current actual channel influence. Then the intermediate conversion vector is binarized by the Sigmoid function into the channel pruning strategy vector $E_l$, and the value of $E_l$ is between 0-1.

In the training process, taking the classification model as an example, we perform pruning layer by layer. We design the loss function so that the model not only pays attention to the classification loss, but also pays attention to the pruning strategy of each layer of the model. We suggest calculating the Euclidean distance between the target encoding vector and the processed pruning strategy vector $E_l$, and use the obtained value as the regular term of the DNN classification model: 
\begin{equation}
    \mathcal{L} = \mathcal{L}_{\text {classification}}+\lambda \| E_l-T_l\|_{2}^{2}
\label{loss}
\end{equation}
In the loss function $\mathcal{L}$, the target encoding vector $T_l$ is used as the teacher to guide the learning of the micro-convolutional layer $K$, so that the micro-convolutional layer can be closer to the target when extracting the characteristics of channel influence and at the same time have generalization. 
This is similar to L1 and L2 regularization, constraining the changing trend of $E_l$. 

Obviously, in order to ensure that the classification loss of the DNN model and the loss proportion of the pruning strategy can be balanced, the loss function needs a hyperparameter $\lambda$. For $\lambda$ we have formulated the following rules: 
\begin{equation}
    \lambda = \left\{
            \begin{array}{lr}
             5.0\times |\frac{T_l}{|C^l|}+\frac{B_l}{|C^l|}-1|, & 1-\frac{B_l}{|C^l|} \geq \frac{T_l}{|C^l|}\\
             0, & \text{otherwise.}
             \end{array}
    \right.
\label{lambda}
\end{equation}
in which $C^l$ represents the original number of channels in the current layer; we recommend that $E_l$ value is set to 0 when the value is less than $10^{-6}$, otherwise it is set to 1, so as to calculate the encoding of the actual binary pruning remaining channel vector $B_l$. 
In the equation \ref{lambda}, $\frac{B_l}{|C^l|}$ represents the actual channel retention rate and $\frac{T_l}{|C^l|}$ represents the actual channel retention rate of the current layer. If the actual channel retention rate rate exceeds the target compression rate, then we should ignore the concern of the model pruning strategy, and should focus on the model's classification (or regression) accuracy. If the actual channel retention rate is lower than the target compression rate, we should appropriately increase the value of $\lambda$ to make the model pay more attention to the realization of the pruning strategy. 
Therefore, when the actual channel retention rate exceeds the target compression rate, we set $\lambda$ to 0, otherwise increase the value of $\lambda$ by a certain multiple.

Note that, for determining the threshold setting of the target pruning strategy vector $T_l$, we first sort the channel influence of the entire model, and then calculate the influence threshold through the compression rate of the entire model. 
This solves the problem of differences in channel influence between different layers. 
In addition, the pruning strategy determined by the threshold obtained by this method may cause all channels in the layer to be pruned. 
In this case, we choose to sort only the channel influence of the current layer, subtract the channels with less influence by sorting, and retain the number of channels 0.2$\times$ the global compression rate.

Finally,  during the training process, we transform the code $E_l$ into the actual binary strategy code vector $B_l$. However, in order to reduce the contingency of the pruning scheme, the application of $B_l$ in the model does not directly reset the weight to 0, but uses the encoding vector $B_l$ as a form of channel mask, and uses the feature and $B_l$ of each channel calculated from the input data as multiplying the channel masks is what we call "false pruning". Pruning and splicing through "false pruning" continuously update the connection weights and set the loss regular term in the loss function until the loss function converges and when the $E_l$ code vector becomes stable, the $E_l$ is converted into a binary code vector $B_l$, and then applied 
Based on the pruning scheme of the current layer of the network, the sparse channel model will be able to produce excellent accuracy.
\subsection{Binarization}
\label{binary}
Since the measurement of parameter importance will affect the state of the network connection, the parameter influence degree is used as the decision index for pruning, and it needs to be mapped to a 0-1 index code vector, which is essential to ensure the quality of pruning. 
Combined with mini-batch pool operations, binarization ensures that all examples in the mini-batch can eventually be converted to the same unique index code. 
We use the scaled Sigmoid mapping function to generate approximate binary code:
\begin{equation}
    B_l = sigmoid(\beta E_l)
\label{bi_eq}
\end{equation}
Where $E_l$ is the data obtained after passing through the micro-convolutional layer, we take $E_l$ as the input and $B_l$ as the output of the mapping function. 
In order to improve the robustness of our method, we introduce a hyperparameter $\beta$ to locally enhance or weaken the input data in advance. 
The $\beta$ value is used as a hyperparameter to control the discrete degree of the input value. 
By gradually increasing the value of $\beta$, the Sigmoid function can be gradually mapped to an approximate binary code close to both ends of 0 and 1.

When the pruning strategy $B_l$ is obtained in each iteration of the model, the channels with less influence will be temporarily pruned, while other channels with greater influence will be retained. 
Obviously, the Sigmoid function has an important influence on the final compression strategy. 
When $\beta$ is small, the model mainly allows micro-convolutional layer to have enough time to learn the parameters. 
After the micro-convolutional layer is fully learned, the $\beta$ value is set to gradually increase, so that the discrete degree of the input value of the equation \ref{bi_eq} gradually increases. 
When $\beta$ is large enough, the approximate binary value will eventually become a value of approximately 0 or 1. 
In other words, the pruning is done during the fine-tuning process, and which filter should be pruned is completely determined by the network itself.

This fine-tuning binarization process helps to obtain a more accurate model. When the influence of some channels becomes smaller $(0.5\rightarrow 0)$, the corresponding filter will gradually stop updating. At the same time, other channels are getting larger and larger $(0.5\rightarrow 1)$, which will force the network to pay more attention to reserved filters.

Another major advantage of the binarization process is that pruning and fine-tuning can now be seamlessly integrated. Pruning can be done during fine-tuning of the model. If the binary code of the channel corresponding to the $B_l$ element is 0, then we know that for any input image, its activation value is always 0. On the contrary, if the binary code of the channel corresponding to the $B_l$ element is 1, the activation value will not change during the element-wise multiplication. Therefore, after the strategy vector $B_l$ converges to binary, deleting all pruned blocks and pruned filters will not change the prediction of the network.

Considering that the loss function during the training of the pruning model is more complicated than the standard cross-entropy loss function, we will take more steps to control its convergence process.

Since $B_l$ is affected by the input tensor, weight value and $\beta$ during training, if the convergence is slow, the possible reason is that the $\beta$ is too small, and $B_l$ may be difficult to converge or even impossible to converge to a binary value. Even $B_l$ may have serious data offset, staying at a small value, that is, all elements are less than 0.5, but it can never be pulled back to about 1. Unfortunately, due to the difference in the input, the appropriate $\beta$ value may vary greatly in different layers. Therefore, it is impossible to find a suitable static $\beta$ for all layers. Based on the above observations, we propose an adaptive scheme to adjust the approximate value of $\beta$.

This can be done by initializing the $\beta$ value first, and then dynamically constraining the discrete degree of input according to the training situation. 
We recommend setting the value of $\beta$ to be small in the initial stage, which can slow down the fitting speed of the pruning strategy. Because $\beta$ is too large will speed up the determination of the pruning strategy, it is likely to cause the pruning strategy to degenerate into a random selection. In fact, before Sigmoid binarization, we use the micro-convolutional layer to learn the pruning strategy to improve its generalization. If the $\beta$ value is too large, the micro-convolutional layer learning is incomplete, which will lead to policy degradation. After initial training with a small $\beta$ value, the convolutional layer has the ability to extract sensitivity generalization features, and gradually increase the $\beta$ value to increase the degree of data dispersion, so that the influence $E_l$ gradually accelerates and converges into a binary number strategy vector $B_l$.

We still solve the convergence problem by separately pruning the convolutional layer and the fully connected layer in the above-mentioned dynamic manner.

\section{Experiments}
\label{experiment}
In this section, we conduct extensive experiments to evaluate our approach on the image classification task, which is the most widely used task for testing pruning methods. 
We introduced our main experimental results in Section \ref{result}. 
In Section \ref{analy}, we discussed the role of hyper-parameters in our method. Finally, we tried a wider FLOP reduction range.
\begin{table*}[]
\centering
\caption{Results on CIFAR-10 and CIFAR-100. Best results are bolded.}
\label{table1}
\begin{tabular}{cccclccc}
\hline
\multirow{2}{*}{Dataset}   & \multirow{2}{*}{Model}     & \multirow{2}{*}{Approach} & Baseline &  & \multicolumn{3}{c}{Pruned}                                                                                                       \\ \cline{4-4} \cline{6-8} 
                           &                            &                           & Acc.(\%) &  & Acc.(\%) & \begin{tabular}[c]{@{}c@{}}Acc. \\ Drop(\%)\end{tabular} & \begin{tabular}[c]{@{}c@{}}FLOPs \\ Reduction\end{tabular} \\ \hline
\multirow{11}{*}{CIFAR-10} & \multirow{8}{*}{ResNet-56} & NS\cite{liu2017learning}                        & 93.80    &  & 93.27    & 0.53                                                     & 48\%                                                       \\
                           &                            & CP\cite{he2017channel}                        & 92.80    &  & 91.80    & 1.00                                                     & 50\%                                                       \\
                           &                            & AMC\cite{he2018amc}                       & 92.80    &  & 91.90    & 0.90                                                     & 50\%                                                       \\
                           &                            & DCP\cite{zhuang2018discrimination}                       & 93.80    &  & 93.49    & 0.31                                                     & 50\%                                                       \\
                           &                            & DCP-adapt\cite{zhuang2018discrimination}                 & 93.80    &  & 93.81    & -0.01                                                    & 47\%                                                       \\
                           &                            & CCP\cite{peng2019collaborative}                       & 93.50    &  & 93.46    & 0.04                                                     & 47\%                                                       \\s
                           &                            & PRNSP\cite{zhuang2020neuron}                     & 93.80    &  & 93.83    & -0.03                                                    & 47\%                                                       \\
                           &                            & Ours                      & 93.89    &  & 94.40    & -0.51                                                    & 41\%                                                           \\ \cline{2-8} 
                           & \multirow{3}{*}{VGG-16}    & NS\cite{he2017channel}                        & 93.58    &  & 93.54    & 0.04                                                     & 34\%                                                       \\
                           &                            & PRNSP\cite{zhuang2020neuron}                     & 93.88    &  & 93.92    & -0.04                                                    & 54\%                                                       \\
                           &                            & Ours                      & 92.11    &  & 92.46    & -0.35                                                    &41\%                                                            \\ \hline
\multirow{6}{*}{CIFAR-100} & \multirow{3}{*}{ResNet-56} & NS\cite{he2017channel}                        & 72.49    &  & 71.40    & 1.09                                                     & 24\%                                                       \\
                           &                            & PRNSP\cite{zhuang2020neuron}                     & 72.49    &  & 72.46    & 0.03                                                     & 25\%                                                       \\
                           &                            & Ours                      & 72.49          &   &  72.51        & -0.02                                                         & 30\%                                                           \\ \cline{2-8} 
                           & \multirow{3}{*}{VGG-16}    & NS\cite{he2017channel}                        & 73.83    &  & 74.20    & -0.37                                                    & 38\%                                                       \\
                           &                            & PRNSP\cite{zhuang2020neuron}                     & 73.83    &  & 74.25    & -0.42                                                    & 43\%                                                       \\
                           &                            & Ours                      & 72.44         &  & 73.13         & -0.69                                                         & 43\%                                                           \\ \hline
\end{tabular}
\end{table*}
\subsection{Datasets and Compared Methods}
\label{setting}
In actual experiments, the effect of the same pruning method on small data sets and large data sets may be different \cite{gale2019state}. Therefore, we evaluated our method on a small dataset and a large dataset. We use CIFAR10 \cite{krizhevsky2009learning} to perform the experiment. In addition, we also tried two widely used deep CNN structures: VGG-Net \cite{simonyan2014very} and ResNet \cite{he2016deep}.

Obviously, it is difficult for us to make direct comparisons with previous works, because many works have only experimented on a data set or on a network structure. 
Therefore, we choose to compare each data set and network structure with only those works that have published experimental results on that data set and network structure. 
We will also run the published code method in an experimental environment. 
If the effect of our operation is better than the original method, the better result will be used in the model comparison, and the suffix of "Our-impl" will be marked on this method in the result table. 
Otherwise, we will directly use the original published results of the published works for comparison. 
We reviewed all the methods that will be compared in Section 2. 
Some methods, such as CCP-AC \cite{peng2019collaborative} use inter-channel dependence to screen channel combinations and require the insertion of additional classifiers. 
As \cite{peng2019collaborative} pointed out, it is unfair to directly compare these methods with other methods. 
Therefore, we will not compare with those pruning methods that require additional classifiers.
\subsection{Experimental setup}
Our code implementation is based on PyTorch and Torchvision \cite{paszke2019pytorch}. And our basic models is based on \cite{simonyan2014very} and \cite{luo2020autopruner} respectively. Moreover, each convolutional layer adds a BN on the basis of the VGG.
We list detailed parameters in the training in the appendix. 
We dynamically adjusted the hyperparameters $\lambda$ and $\beta$ during the training process to control the reduced FLOPs. For ResNet, we only prune blocks with residual connections, and the detailed information of pruning ResNet50 is shown in the appendix.
\subsection{Experiment Results}
\label{result}

\subsection{Analysis}
\begin{table*}[]
\centering
\caption{Results on CIFAR-10 and CIFAR-100. Best results are bolded.}
\label{table2}
\begin{tabular}{ccccccc}
\hline
\multirow{2}{*}{Dataset/Model}    & \multirow{2}{*}{Approach} & Baseline & \multirow{2}{*}{} & \multicolumn{3}{c}{Pruned}                                                                                                     \\ \cline{3-3} \cline{5-7} 
                                  &                           & Acc.(\%) &                   & Acc.(\%) & \begin{tabular}[c]{@{}c@{}}Acc.\\ Drop(\%)\end{tabular} & \begin{tabular}[c]{@{}c@{}}FLOPs\\ Reduction\end{tabular} \\ \hline
\multirow{3}{*}{CIFAR10/Resnet56} & NS                        & 93.80    &                   & 91.20    & 2.60                                                    & 68\%                                                      \\
                                  & PRNSP                     & 93.80    &                   & 92.63    & 1.17                                                    & 70\%                                                      \\
                                  & Ours                      & 93.89    &                   & 94.40    & -0.51                                                   & 41\%                                                      \\ \hline
\end{tabular}
\end{table*}
\label{analy}
\textbf{Compression rate $r$}: In our work, the target compression rate $r$ represents the target compression rate of the channel, which is the ratio of the number of pruned channels to the total number of channels. We take the compression rate as a predefined value and it is the target of filter pruning. However, the final compression rate of the model may be slightly smaller than the target compression rate $r$. The reason is that our pruning method is adaptive. We add a regularizer to the loss function to encourage the model to pay attention to the compression rate of the model, so that the model can trade-off between accuracy and parameter compression rate. The accuracy achieved by the actual compression rate is shown in Figure xx, and the actual compression accuracy changes under different compression rates can be seen. Figure xx shows that when the target compression ratio $r$ is a certain value, the actual compression ratio is always slightly lower than the target value. This is reasonable. We added the $\lambda$ hyperparameter to the loss function to adjust the ratio of the classification loss to the pruning strategy loss. When the actual compression rate is less than the target compression rate, the $\lambda$ value changes according to the equation \ref{bi_eq}. The closer the actual compression rate is to the target compression rate, the smaller the value of $\lambda$. Therefore, $\lambda$ encourages model training to pay attention to the accuracy and compression rate of the model at the same time. The larger the value of $\lambda$, the more attention is paid to the compression strategy of the model during training, and vice versa, the more attention is paid to the classification loss. When the actual compression rate exceeds the target compression rate, we believe that the compression rate has reached the requirement. At this time, the $\lambda$ is set to $0$. At this time, the model only pays attention to the accuracy change of the model. In the training process, the model needs to determine the degree of attention to the pruning strategy based on $\lambda$ value, and from \cite{luo2020autopruner}, it can be seen that the $\lambda$ change mechanism is determined by the model itself, so the actual pruning rate is adaptive.

\textbf{The parameter $\beta$}: As Eq. (\ref{bi_eq}) shows that $\beta$ is an important hyperparameter. And we only need to pay attention to $\beta_{start}$. $\beta_{stop}$ can be adjusted jointly according to the binarization situation and the number of training times during model training. $\beta_{start}$ cannot be too large. If $\beta_{start}$ is too large, the value of the $Sigmoid$ mapping is completely binarized in advance, which will make the convolutional layer of the learning model strategy unable to fully learn the features of the influence of the model parameters. In this case, our method will degenerate into random channel selection. Different models have different $\beta_{start}$ to ensure sufficient model learning, so $\beta_{start}$ is related to the network model. We have studied the influence of hyperparameter $\beta_{start}$ on the accuracy of the model based on experience. Figure xx shows that when the $\beta_{start}$ is large, the influence of the model weight cannot be fully learned, so the accuracy is slightly lower. When $\beta_{start}$ is at a small value ($\beta_{start}=0.01$), the prediction accuracy reaches the highest. But when $\beta_{start}$ is small to a certain extent, the prediction accuracy of the model drops instead. The reason is that the $\beta_{start}$ is too small, and the over-learning of the convolutional layer of the learning model strategy leads to a slight overfitting of the model, which leads to a decrease in model prediction accuracy. The results in Figure xx show that our method needs to choose a small $\beta_{start}$ (within a certain range), and the previous few trainings can be kept small to achieve better results. At initialization, we set $\beta_{end}$ to be 100 times larger than $\beta_{start}$.

\textbf{More extensive experiments to reduce FLOPs}:
For the channel trim work we need to be FLOPs reduce the amount of broad-spectrum test. Our Flops calculation method is the same as that of \cite{zhuang2020neuron}. We compare it with other channel pruning methods: the PRNSP\cite{zhuang2020neuron}, which uses $P$-regularization to make the BN layer only reduce the scale coefficient of the "unimportant" channel, and keep other coefficients still in a larger state. We also compare NS\cite{liu2017learning}. For a fair comparison, we use the same experimental setup to run all methods in this article, which means that all methods use exactly the same number of training epochs. The results are listed in Table \ref{table2}, which shows that our method consistently shows better performance in the case of extensive FLOPs reduction.
\section{Conclusion}

In this paper, we propose a new reliable method for measuring the influence of weights. We use the influence function to calculate the influence, and pruning is achieved by cutting off the channels with less influence. We theoretically prove that the back propagation of the deep network is the first-order Taylor approximation of the influence function of the weight. And through the experimental method to eliminate the influence of different data on the measurement, thus verifying the effectiveness of the influence function to calculate the influence, and the effectiveness of channel pruning through the magnitude of the influence.
We integrate filter selection and model fine-tuning into a single end-to-end trainable framework. The two processing steps promote each other, the method is easy to implement and computationally efficient.
In the future, we will find more applications that require the influence of parameters.

{\small
\bibliographystyle{ieee_fullname}
\bibliography{egbib}
}

\end{document}